# DeepCloak: Masking Deep Neural Network Models for Robustness Against Adversarial Samples

**Ji Gao[1], Beilun Wang[1], Zeming Lin[1], Weilin Xu[1], Yanjun Qi[1]**
[1] Department of Computer Science
University of Virginia
Charlottesville, VA 22904-4740
`{jg6yd,bw4mw,xuweilin,yanjun}@virginia.edu`

## Abstract

Recent studies have shown that deep neural networks (DNN) are vulnerable to adversarial samples: maliciously-perturbed samples crafted to yield incorrect model outputs. Such attacks can severely undermine DNN systems, particularly in security-sensitive settings. It was observed that an adversary could easily generate adversarial samples by making a small perturbation on irrelevant feature dimensions that are unnecessary for the current classification task. To overcome this problem, we introduce a defensive mechanism called DeepCloak. By identifying and removing unnecessary features in a DNN model, DeepCloak limits the capacity an attacker can use generating adversarial samples and therefore increase the robustness against such inputs. Comparing with other defensive approaches, DeepCloak is easy to implement and computationally efficient. Experimental results show that DeepCloak can increase the performance of DNN models against adversarial samples.

## 1 Introduction

Deep Neural Networks (DNNs) have achieved great success in a variety of applications. Classifiers based on DNN models have attained great performance on multiple security-sensitive tasks (Microsoft Corporation, 2015; Dahl et al., 2013). However, recent studies show that machine learning classifiers are vulnerable to deliberate attacks. Attackers can easily generate a malicious sample by adding a small perturbation to a normal sample. Then the malicious sample can fool the classifier and force it to yield a wrong output. Such sample is called an *adversarial sample*. This paper focuses on finding a defensive approach that can make DNN models perform more reliable in the face of adversarial samples.

Many recent studies focused on adversarial samples, including Szegedy et al. (2013); Papernot et al. (2015); Goodfellow et al. (2014); Wang et al. (2016); Papernot et al. (2016b;a). Different algorithms for generating adversarial samples have been invented. Szegedy et al. (2013) proposed an algorithm to generate adversarial samples using Box L-BFGS method. It also showed that same adversarial sample could be transferred to fool different DNN classifiers. Goodfellow et al. (2014) purposed *fast gradient sign method*, which maximizes the consequence of the attack under limited size of $L_\infty$-norm. Papernot et al. (2015) purposed another algorithm that generates adversarial samples following the saliency value. A recent paper Papernot et al. (2016a) proposed a black-box attack, which first approximates the target classifier and then generates adversarial samples to the approximated model.

Researchers also studied how to defend attacks enforced by adversarial samples. Goodfellow et al. (2014) showed retraining a new model using adversarial samples can improve the adversarial robustness of the model. Papernot et al. (2016b) proposed *defensive distillation* to make the model less sensible to gradient change. Wang et al. (2016) showed that one cause of adversarial sample is those redundant features that are unnecessary to classification. A perturbation along an unnecessary feature dimension can easily fool a classifier.

In this paper, we introduce a new defensive approach for DNN models: DeepCloak. The motivation of DeepCloak is to increase model robustness by removing unnecessary features. Comparing to previous defensive approach, DeepCloak has the following benefits: 1. DeepCloak doesn't need additional training process. Therefore it is computationally efficient. 2. DeepCloak can be easily applied to any base DNN models.

In the rest part, Section 2 briefly introduces the background of adversarial samples and its relationship to unnecessary features. Section 3 introduces our defense approach DeepCloak. It removes unnecessary features in trained DNNs and thus increase the adversarial robustness. In Section 4, experiment results show that DeepCloak works effectively for a popular DNN model.

## 2 BACKGROUND

### 2.1 ADVERSARIAL SAMPLES

For the following analysis, we denote a base DNN classifier as $F(\cdot) : \mathbb{R}^n \to \mathbb{Y}$, where $x \in \mathbb{R}^n$ denotes a single sample and $\mathbb{Y}$ is the set of output classes.

Adversarial samples are deliberately created samples. We define of an adversarial sample $x'$ as:
$$x' = x + \Delta x, |\Delta x| < \epsilon, x' \in \mathbb{X}$$
$$F(x) \neq F(x') \quad (2.1)$$

We assume inputs close to each other are similar. For example, in image classification, $\Delta x$ needs to be small so that $x$ and $x'$ are imperceptible to human eyes while the machine classifier $F$ still classifies $x$ and $x'$ into two different classes.

### 2.2 ADVERSARIAL SAMPLES AND UNNECESSARY FEATURES

Wang et al. (2016) show that the effectiveness of adversarial samples is related to extra unnecessary features extracted by the machine classifier. If a classifier extracts many unnecessary features in the training process, it will be vulnerable to adversarial attacks.

To solve this problem, we then propose an approach that can reduce the number of unnecessary features. Presented in Figure 2 of Appendix Section 6.2, the basic motivation is that the distance between an adversarial sample and its seed example will be small along *relevant* feature dimensions and relatively large along the *unnecessary* dimensions for the current task. Therefore, we propose to remove unnecessary features for a DNN model by directly modifying its network structure.

## 3 METHOD: DEEPCLOAK

The basic idea of DeepCloak is to remove unnecessary features that can be used for generating adversarial samples. To identify which feature is unnecessary, we test pairs of adversarial samples $x'$ and its normal seed $x$, and compare the difference between the extracted features in DNN. Those features changed rapidly are utilized by the adversary, and thus should be removed to improve the robustness of the model.

To remove those unnecessary features, we insert a mask layer in a DNN model right before the linear layer handling classification. Th mask layer serves as a selector, which will keep the necessary features and remove the unnecessary features by setting them to 0. The size of the feature vector is kept unchanged. The proposed structure is shown in Figure 1.

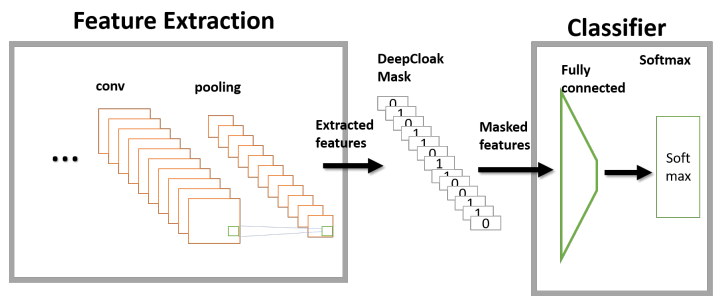

**Figure 1: A sketch of DeepCloak: A mask layer with weights either 0 or 1 is added right before the classification layers.**

The input of the mask layer is the feature vector extracted by previous layers of the DNN model. The weight of the mask layer is either 0 or 1. An element-wise multiplication is done by the mask layer. Therefore, the output is either the input feature or 0. We remove the top $n\%$ of features with highest sensitivity to adversarial samples.

The process is summarized in Algorithm 1. $X$ can be a subset of the full training set, so the process of learning the mask can be fast. No retraining of the model is needed.

**Algorithm 1** DeepCloak algorithm

**Input:** Training set $X = \{x_1, x_2 \ldots x_N\}$, DNN classifier $F()$, adversarial power $\epsilon$. $g()$ represents feature extraction layers of $F()$
1: Initialize a vector $v$ with all 0.
2: **for** $i = 1, 2, \ldots, N$ **do**
3:     Generate an adversarial sample $x'_i$ using sample $x_i$ with power $\epsilon$.
4:     Forward $x_i$ into the network, get the output feature vector $g(x_i)$
5:     Forward $x'_i$ into the network, get the output feature vector $g(x'_i)$
6:     Add $|g(x_i) - g(x'_i)|$ into $v$.
7: **end for**
8: Set $v$'s top $m$ entries to 0 and the rest to 1.
9: Insert the mask layer $v$ into DNN $F$ after the inspected feature layer.

Essentially the sensitivity of are accumulated into the v vector as $v = \sum_{i=1}^{N} |g(x_i) - g(x'_i)|$ after step 7 of Algorithm 1.

## 4 EXPERIMENTS

### 4.1 EXPERIMENT SETTING

- Dataset: We choose CIFAR-10 (Krizhevsky & Hinton, 2009), an image dataset with 50,000 32x32 training images and 10,000 testing images.

- We choose a Residual network with 164 layers(He et al., 2016) as our target DNN model. The model is pre-trained and achieves high performance on the corresponding dataset.

- Metric: We generate adversarial samples for every sample in the test set and test all adversarial samples on each DNN model. The accuracy on the adversarial sample set is reported as "adversarial accuracy" to measure the adversarial robustness of a DNN model.

### 4.2 EXPERIMENT RESULT

The result of DeepCloak against adversarial samples is displayed in Table 1. DeepCloak can reduce the effectiveness of such adversarial samples by $10\%$. The adversarial perturbation is generated using fast gradient sign method using $\epsilon = 10$ (Appendix Section 6.1).

| Nodes masked(%) | Adversarial accuracy | Relative increase | Accuracy | Relative decrease |
|---|---|---|---|---|
| 0% | 0.2961 | 0.00% | 0.943 | 0.00% |
| 1% | 0.3923 | 32.49% | 0.9372 | -0.62% |
| 2% | 0.4234 | 42.99% | 0.9132 | -3.16% |
| 3% | 0.414 | 39.82% | 0.9093 | -3.57% |
| 4% | 0.4146 | 40.02% | 0.8954 | -5.05% |
| 5% | 0.4229 | 42.82% | 0.9 | -4.56% |
| 6% | 0.4173 | 40.93% | 0.9017 | -4.38% |

**Table 1: The result of DeepCloak on Res-net against adversarial attacks. There are totally 256 nodes in the feature output layer.**

Table 1 also indicates that only a small percentage of features are important for adversarial samples. In the table, masking $1\%$ of features can increase the performance by $10\%$ in the adversarial setting. Therefore, we only need to remove a small percent of features to improve the adversarial robustness greatly, and the model still achieves high accuracy on normal test samples.

## 5 CONCLUSION

In this study, we present DeepCloak, a simple and cost-efficient strategy to reduce the effectiveness of adversarial samples on DNN classifiers. Experiments shows that DeepCloak can increase model performance in the adversarial setting. In the future, we will explore more approaches that can reduce the number of feature dimensions to increase the adversarial robustness.

## REFERENCES

George E Dahl, Jack W Stokes, Li Deng, and Dong Yu. Large-scale malware classification using random projections and neural networks. In *ICASSP*, 2013.

Ian J Goodfellow, Jonathon Shlens, and Christian Szegedy. Explaining and harnessing adversarial examples. *arXiv preprint arXiv:1412.6572*, 2014.

Kaiming He, Xiangyu Zhang, Shaoqing Ren, and Jian Sun. Deep residual learning for image recognition. In *Proceedings of the IEEE Conference on Computer Vision and Pattern Recognition*, pp. 770–778, 2016.

Alex Krizhevsky and Geoffrey Hinton. Learning multiple layers of features from tiny images. *Technique report, University of Toronto*, 2009.

Yann LeCun, Léon Bottou, Yoshua Bengio, and Patrick Haffner. Gradient-based learning applied to document recognition. *Proceedings of the IEEE*, 86(11):2278–2324, 1998.

Microsoft Corporation. Microsoft Malware Competition Challenge. https://www.kaggle.com/c/malware-classification, 2015.

Nicolas Papernot, Patrick McDaniel, Somesh Jha, Matt Fredrikson, Z Berkay Celik, and Ananthram Swami. The limitations of deep learning in adversarial settings. *arXiv preprint arXiv:1511.07528*, 2015.

Nicolas Papernot, Patrick McDaniel, Ian Goodfellow, Somesh Jha, Z Berkay Celik, and Ananthram Swami. Practical black-box attacks against deep learning systems using adversarial examples. *arXiv preprint arXiv:1602.02697*, 2016a.

Nicolas Papernot, Patrick McDaniel, Xi Wu, Somesh Jha, and Ananthram Swami. Distillation as a defense to adversarial perturbations against deep neural networks. In *Security and Privacy (SP), 2016 IEEE Symposium on*, pp. 582–597. IEEE, 2016b.

Karen Simonyan and Andrew Zisserman. Very deep convolutional networks for large-scale image recognition. *arXiv preprint arXiv:1409.1556*, 2014.

Christian Szegedy, Wojciech Zaremba, Ilya Sutskever, Joan Bruna, Dumitru Erhan, Ian Goodfellow, and Rob Fergus. Intriguing properties of neural networks. *arXiv preprint arXiv:1312.6199*, 2013. URL http://arxiv.org/abs/1312.6199.

Beilun Wang, Ji Gao, and Yanjun Qi. A theoretical framework for robustness of (deep) classifiers under adversarial noise. *arXiv preprint arXiv:1612.00334*, 2016.

Sergey Zagoruyko and Nikos Komodakis. Wide residual networks. *arXiv preprint arXiv:1605.07146*, 2016.

## 6 APPENDIX

### 6.1 FAST GRADIENT SIGN ALGORITHM

As introduced in Section 1, multiple algorithms have been proposed to generate the adversarial sample $x$. Fast Gradient Sign Algorithm proposed by Goodfellow et al. (2014) is one efficient algorithm in creating adversarial samples. The perturbation is calculated through the following equation:

$$\Delta x = \epsilon \, \text{sign}(\nabla_z Loss(F(z), F(x))) \tag{6.1}$$

This is motivated by controlling the $L_\infty$ norm of $\Delta x$, which results in the size of perturbation in each feature dimension are the same. Maximizing the difference between $F(x)$ and $F(x')$ while limiting the $L_\infty$ norm of $\Delta x$ means to follow the gradient direction on every dimension, which is exactly the sign of the gradient.

### 6.2 EXAMPLE OF AN UNNECESSARY FEATURE CAN BE UTILIZED BY ADVERSARIAL SAMPLES

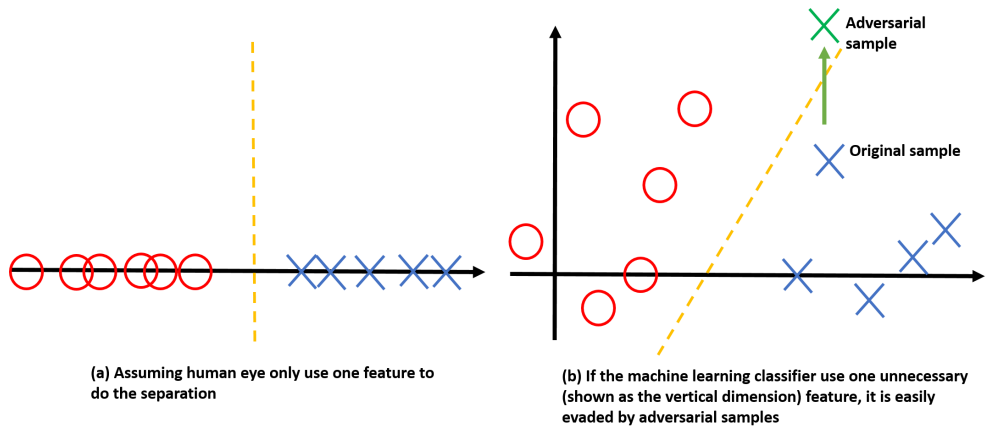

(a) Assuming human eye only use one feature to do the separation

(b) If the machine learning classifier use one unnecessary (shown as the vertical dimension) feature, it is easily evaded by adversarial samples

**Figure 2: One possible type of adversarial vulnerability when learning a linear classifier from unnecessary features**

Figure 2 shows a situation when a linear classification model uses an unnecessary feature. For simplicity, we assume human annotators only use one feature to do the classification. Since the machine classifier use one extra feature, the attacker can push the original sample (blue cross) along that extra feature dimension (y-axis) to generate an adversarial samples (green cross), which is misclassified by the machine classification model.

### 6.3 MORE EXPERIMENTS

#### 6.3.1 EXPERIMENT ON DIFFERENT MODELS AND DATASETS

We add experiments to show our method can be applied to a wide range of DNN models. The results are displayed in Table 2, Table 3 and Table 4.

More specifically, we train a small CNN on the MNIST dataset (LeCun et al., 1998) , and also VGG (Simonyan & Zisserman, 2014) model and Wide Residual Network model (Zagoruyko & Komodakis, 2016) on the CIFAR-10 dataset.

| Nodes masked(%) | Adversarial accuracy | Relative increase | Accuracy | Relative decrease |
|---|---|---|---|---|
| 0% | 0.5586 | 0.00% | 0.9827 | 0.00% |
| 1% | 0.5604 | 0.32% | 0.9821 | -0.06% |
| 2% | 0.5589 | 0.05% | 0.9817 | -0.10% |
| 3% | 0.5573 | -0.23% | 0.9817 | -0.10% |
| 4% | 0.5601 | 0.27% | 0.9820 | -0.07% |
| 5% | 0.5644 | 1.04% | 0.9816 | -0.11% |
| 6% | 0.5691 | 1.88% | 0.9807 | -0.20% |
| 7% | 0.5791 | 3.67% | 0.9812 | -0.15% |
| 8% | 0.5794 | 3.72% | 0.9810 | -0.17% |
| 9% | 0.5802 | 3.87% | 0.9808 | -0.19% |
| 10% | 0.5839 | 4.53% | 0.9807 | -0.20% |

Table 2: The result of DeepCloak on a small CNN model(with 2 conv layers, 2 pooling layers and a linera layer) trained on MNIST against adversarial samples. There are totally 200 nodes in the feature output layer.

| Nodes masked(%) | Adversarial accuracy | Relative increase | Accuracy | Relative decrease |
|---|---|---|---|---|
| 0% | 0.2360 | 0.00% | 0.9363 | 0.00% |
| 1% | 0.2443 | 3.52% | 0.9355 | -0.09% |
| 2% | 0.2481 | 5.13% | 0.9369 | 0.06% |
| 3% | 0.2490 | 5.51% | 0.9363 | 0.00% |
| 4% | 0.2527 | 7.08% | 0.9361 | -0.02% |
| 5% | 0.2539 | 7.58% | 0.9358 | -0.05% |
| 6% | 0.2564 | 8.64% | 0.9354 | -0.10% |
| 7% | 0.2583 | 9.45% | 0.9350 | -0.14% |
| 8% | 0.2609 | 10.55% | 0.9360 | -0.03% |
| 9% | 0.2604 | 10.34% | 0.9346 | -0.18% |
| 10% | 0.2632 | 11.53% | 0.9340 | -0.25% |

Table 3: The result of DeepCloak on VGG network trained on CIFAR-10 against adversarial samples. There are totally 512 nodes in the feature output layer.

| Nodes masked(%) | Adversarial accuracy | Relative increase | Accuracy | Relative decrease |
|---|---|---|---|---|
| 0% | 0.2546 | 0.00% | 0.9537 | 0.00% |
| 1% | 0.2652 | 4.16% | 0.9529 | -0.08% |
| 2% | 0.2683 | 5.38% | 0.9517 | -0.21% |
| 3% | 0.2753 | 8.13% | 0.9512 | -0.26% |
| 4% | 0.2797 | 9.86% | 0.9500 | -0.39% |
| 5% | 0.2828 | 11.08% | 0.9491 | -0.48% |
| 6% | 0.2900 | 13.90% | 0.9486 | -0.53% |
| 7% | 0.2927 | 14.96% | 0.9476 | -0.64% |
| 8% | 0.2963 | 16.38% | 0.9461 | -0.80% |
| 9% | 0.2993 | 17.56% | 0.9470 | -0.70% |
| 10% | 0.2966 | 16.50% | 0.9458 | -0.83% |

Table 4: The result of DeepCloak on Wide Residual Network trained on CIFAR-10 against adversarial samples. There are totally 640 nodes in the feature output layer.

In all three tables, masking a small amount of nodes in the output features can increase the adversarial performance of the model. The accuracy on normal samples is slightly sacrificed. But comparing with the increase of adversarial accuracy, the changes are much smaller. We observe that many features are unimportant for the small CNN and VGG models, as removing 10% of the features lead to a tiny loss in relative accuracy. The percentage of nodes masked should be tuned for different architectures.

#### 6.3.2 COMPARING TO A RANDOM MASK

To justify our result, we compare DeepCloak mask with a random mask. That is, we insert a fixed mask layer with weights 0 and 1 randomly signed. To compare with DeepCloak, same percentage of nodes have been masked. The result has been displayed in Figure 3.

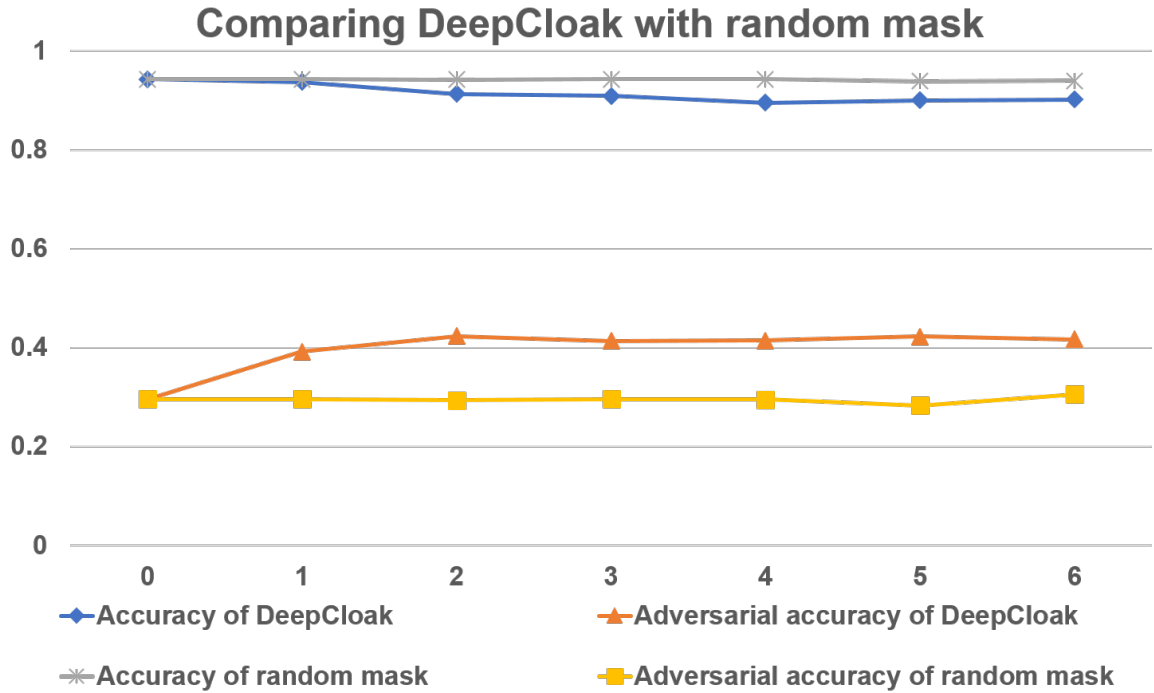

Figure 3: Compare DeepCloak with a random mask

6.3.3 EXPERIMENT OF CHANGING ATTACK POWERS

Experimental result shown in Table 1 to Table 4 fix the attack power ($\epsilon$ of fast gradient sign method) as 10. Here we show the experimental result when changing the value of attack powers.

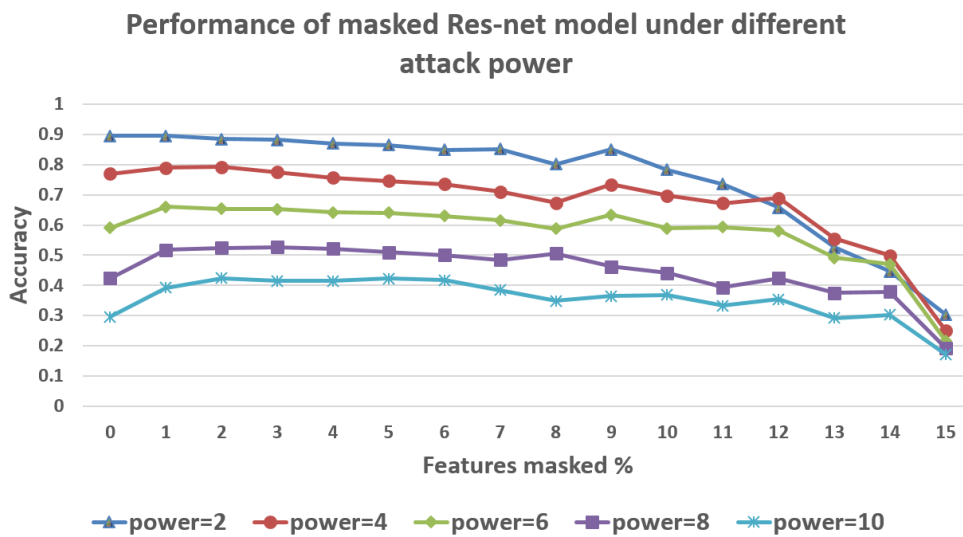

Figure 4: Comparing the performance of DeepCloak when varying attack powers on Res-net. Different colors indicate different attacking powers.

Figure 4 show the experiment result when changing attack powers. The performance of the base model gets better when the attack power becomes smaller. DeepCloak works better when the attack power is large when the original model is highly affected by adversarial samples. For most curves, masking 1% of nodes can lead to better adversarial accuracy already.

### 6.3.4 EXPERIMENT OF MASKING DIFFERENT LABELS SEPARATELY

It is possible that some features only work for some particular output labels. Therefore, we then try to learn different masks for different labels. The process is still like Algorithm 1, but we do it separately for every possible output labels. We remove the same number of features for every labels.

The result is shown in Figure 5. Comparing to only hold a global mask, this method have a slight decrease in performance when the number of features being masked is small. This is probably because different masks make the model biased to some specific labels.

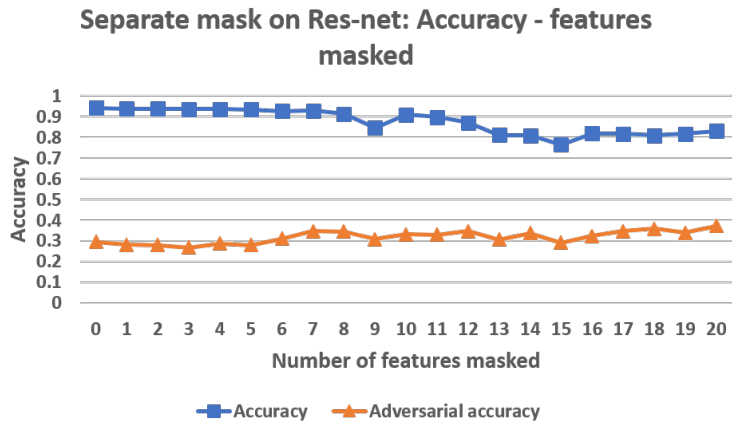

Figure 5: The performance of keeping separate weights